\documentclass[letterpaper]{article} 
\usepackage[]{aaai2026}  
\usepackage{times}  
\usepackage{helvet}  
\usepackage{courier}  
\usepackage[hyphens]{url}  
\usepackage{graphicx} 
\usepackage{natbib}  
\usepackage{caption} 
\usepackage{algorithm}

\usepackage{newfloat}
\usepackage{listings}
\DeclareCaptionStyle{ruled}{labelfont=normalfont,labelsep=colon,strut=off} 
\floatstyle{ruled}
\newfloat{listing}{tb}{lst}{}
\floatname{listing}{Listing}
\usepackage{amsthm}
\usepackage{amsmath}
\usepackage{amsfonts}
\usepackage{amssymb}
\usepackage{booktabs}
\usepackage{multirow}
\usepackage{graphicx}
\usepackage{listings}
\usepackage{tcolorbox}
\tcbuselibrary{listings, breakable, skins}  
\usepackage{amsmath}
\usepackage[noend]{algpseudocode} 

\title{Wisdom of the Crowd:  Reinforcement Learning from \\ Coevolutionary Collective Feedback}
\author{
    Wenzhen Yuan $^{1,2}$\equalcontrib\thanks{Work was done during an internship at Shanghai AI Lab.},
    Shengji Tang $^{2,3}$\equalcontrib,
    Weihao Lin $^{4}$,
    Jiacheng Ruan $^{1}$,
    Ganqu Cui $^{2}$,
    Bo Zhang $^{2}$,
    Tao Chen $^{4}$,
    Ting Liu $^{1}$,
    Yuzhuo Fu $^{1}$\thanks{Corresponding author.},
    Peng Ye $^{2,3\ddagger}$,
    LEI BAI $^{2}$
}
\affiliations{

    $^{1}$ Shanghai Jiao Tong University\\
    $^{2}$ Shanghai Artificial Intelligence Laboratory\\
    $^{3}$ The Chinese University of Hong Kong\\
    $^{4}$ Fudan University\\
    winston\_yuan@sjtu.edu.cn
}

\usepackage{bibentry}

\nocopyright
\begin{document}

\maketitle

\begin{abstract}

Reinforcement learning (RL) has significantly enhanced the reasoning capabilities of large language models (LLMs), but its reliance on expensive human-labeled data or complex reward models severely limits scalability. While existing self-feedback methods aim to address this problem, they are constrained by the capabilities of a single model, which can lead to overconfidence in incorrect answers, reward hacking, and even training collapse.
To this end, we propose Reinforcement Learning from Coevolutionary Collective Feedback (RLCCF), a novel RL framework that enables multi-model collaborative evolution
without external supervision.
Specifically, RLCCF optimizes the ability of a model collective by maximizing its Collective Consistency (CC), which jointly trains a diverse ensemble of LLMs and provides reward signals by voting on collective outputs.
Moreover, each model's vote is weighted by its Self-Consistency (SC) score, ensuring that more confident models contribute more to the collective decision. 
Benefiting from the diverse output distributions and complementary  abilities of multiple LLMs, RLCCF
enables the model collective to continuously enhance its  reasoning ability through coevolution. 
Experiments on four mainstream open-source LLMs across four mathematical reasoning benchmarks demonstrate that our framework yields significant performance gains, achieving an average relative improvement of 16.72\% in accuracy.
Notably, 
RLCCF not only improves the performance of individual models but also enhances the group's majority-voting accuracy by 4.51\%, demonstrating its ability to extend the collective capability boundary of the model collective.

\end{abstract}

\section{Introduction}
The reasoning capabilities of large language models (LLMs) have been significantly enhanced by the advances in reinforcement learning (RL)~\cite{DeepSeekAI2025DeepSeekR1IR, qwen3technicalreport, Contributors2024OpenAIOS}. Early approaches primarily focused on Reinforcement Learning from Human Feedback (RLHF)~\cite{ouyang2022training}, where a reward model is trained to estimate human preferences and guide the model toward generating human-aligned outputs. However, RLHF is bottlenecked by the limited capacity of reward models and the high cost of laborious human annotations. To address this, Reinforcement Learning with Verifiable Rewards (RLVR)~\cite{DeepSeekAI2025DeepSeekR1IR,qwq32b} uses rule-based or automated verifiers to provide reward signals, excelling in structured domains like mathematics and programming. Nevertheless, RLVR still relies heavily on manually designed verifiers and annotated data, limiting its scalability.

\begin{figure*}[!t]
\centering
\includegraphics[width=.45\linewidth]{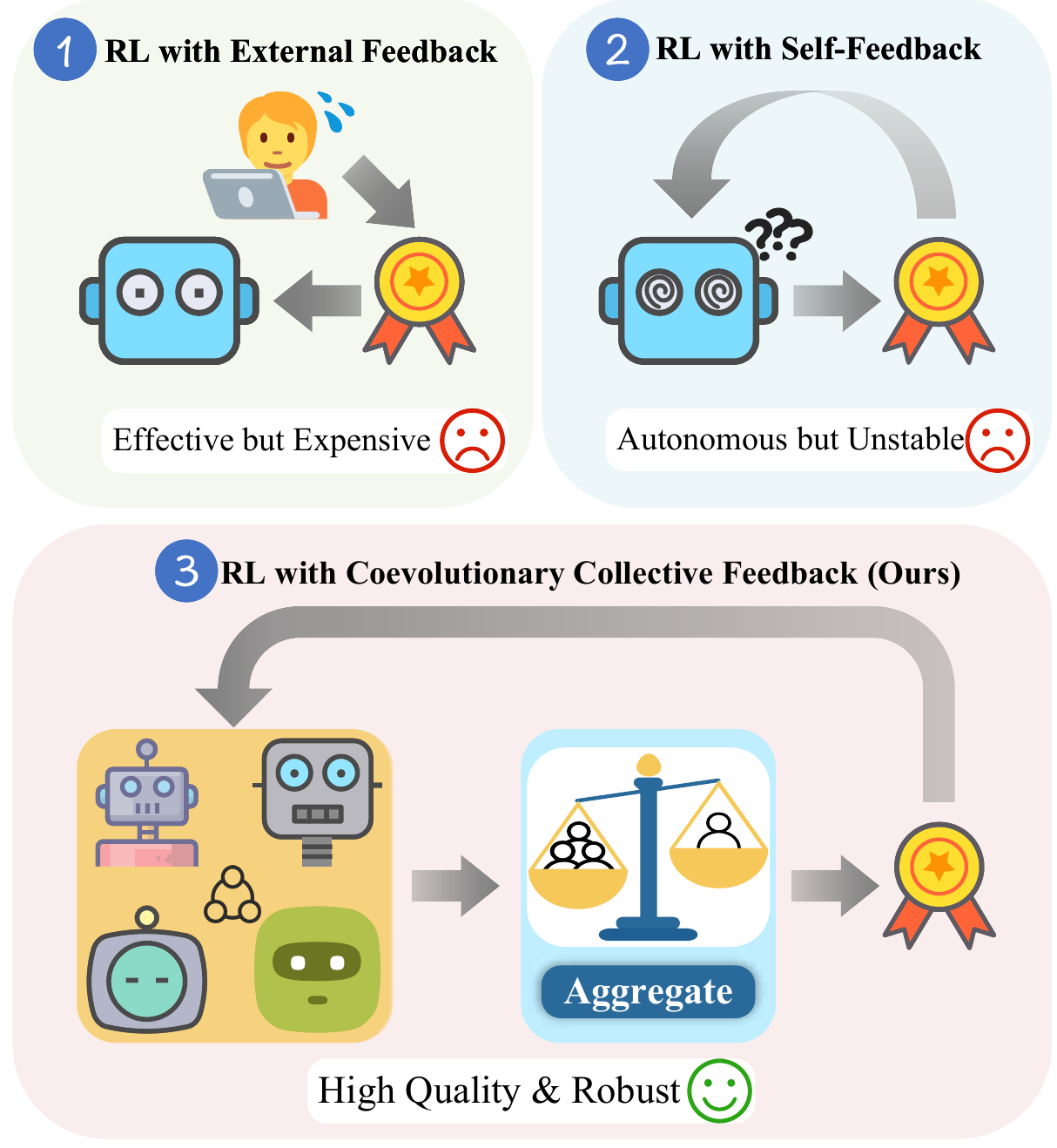}~~
\includegraphics[width=.45\linewidth]{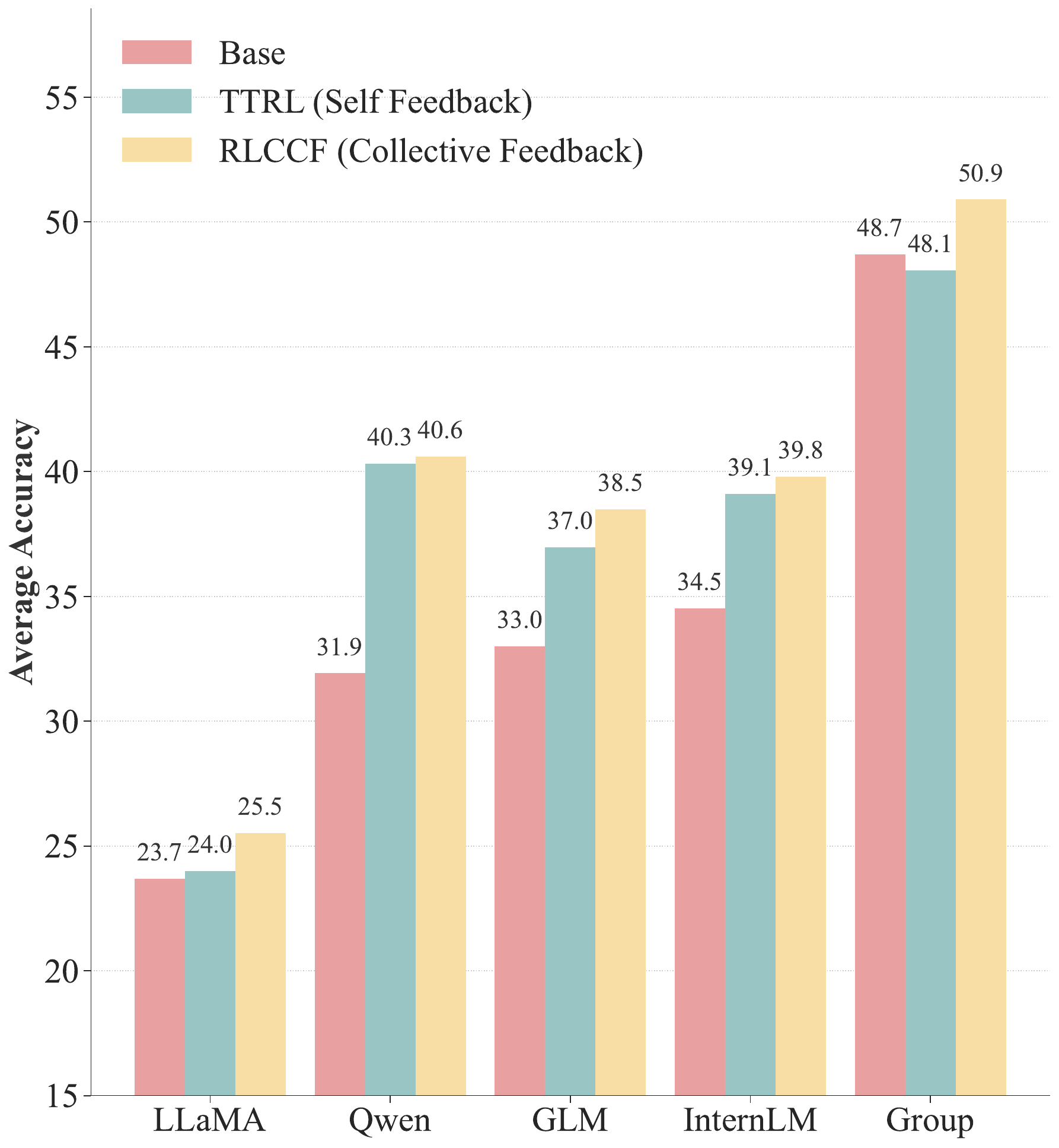}
\caption{Left: Comparison of different RL paradigms for LLMs.
Right: The average accuracy of Qwen2.5-7B, GLM-4-9B, InternLM3-8b-Instruct and LLaMA-3.1-8B-Instruct on four Mathematical benchmarks, ``Group" denotes the majority vote accuracy of all four LLMs originating from different methods. 
}
\label{fig:schematic}
\end{figure*}
To overcome the reliance on external supervision, emerging works have explored RL training paradigms based on self-feedback~\cite{Huang2022LargeLM,Zhao2025AbsoluteZR,Jiao2024PreferenceOF}. 
Methods like TTRL~\cite{Zuo2025TTRLTR} and EMPO~\cite{Zhang2025RightQI} compute rewards from the consistency among the model's multiple samples. Intuitor~\cite{Zhao2025LearningTR}, on the other hand, provides a internal perspective by utilizing the model's own self-certainty as an intrinsic reward signal. 
While these methods remove the need for human annotations, they are constrained by the capabilities of a single model. This limitation might leads to overconfidence in incorrect outputs, and the single-model self-training paradigm can create uncontrolled positive feedback loops, potentially resulting in reward hacking or training collapse~\cite{shumailov2024ai, Zhang2025RightQI}.


In this paper, we draw inspiration from multi-agent systems ~\cite{Chen2025SymbolicMA, Lu2023RoutingTT} and propose Reinforcement Learning from Coevolutionary Collective Feedback (RLCCF), a collaborative and unsupervised RL framework that addresses these issues. RLCCF optimizes a diverse set of LLMs (e.g., models from different developers) by maximizing their Collective Consistency (CC). The process begins with each model independently sampling multiple candidate answers. Then, a Self-Consistency (SC) score is computed for each model's outputs, serving as a robust proxy for internal confidence. These SC scores are used as dynamic weights in majority voting to generate high-quality pseudo-labels, synthesizing the collective knowledge.
The pseudo-label then serves as the supervision signal for all participating models.
Through this process, RLCCF enables the model collective to progressively enhance its Collective Consistency and overall reasoning ability.

We validate RLCCF on four representative open-source LLMs across four challenging mathematical reasoning benchmarks: AIME 2024~\cite{li2024numinamath}, AMC~\cite{li2024numinamath}, MATH-500~\cite{hendrycks2021measuring}, and OlympiadBench~\cite{he2024olympiadbench}. Using only 700 unlabeled train data, RLCCF consistently yields significant improvements. Notably, we observe average gains of 27.20\% for Qwen2.5-7B, 16.63\% for GLM-4-9B, 15.31\% for InternLM3-8b-Instruct, and 7.74\% for LLaMA-3.1-8B-Instruct. Crucially, our framework also boosts the group’s majority-voting accuracy by 4.51\%, a feat not achieved by existing single-model self-feedback methods. This highlights that RLCCF not only makes each model smarter but also effectively extends the collective capability boundary of the entire model group.
Our contributions are summarized as follows:
    
\begin{itemize}
    \item We propose RLCCF, a novel unsupervised RL framework that leverages multi-model coevolution to  resolve the reward hacking and training collapse issues inherent in recent single-model self-feedback methods.
    \item Our experiments demonstrate that RLCCF not only enhances the performance of individual models but makes a breakthrough in extending the collective capability boundary of the entire model group.
    \item We conduct both theoretical and empirical analysis to elucidate the mechanisms of RLCCF from two perspectives: bias reduction and model complementarity.   
\end{itemize}

\section{Related Work}

\subsection{RLHF and RLVR}
Reinforcement learning (RL) has achieved remarkable success in boosting the reasoning capabilities of LLMs~\cite{Contributors2024OpenAIOS, Ouyang2022TrainingLM}. A foundational approach in this area is Reinforcement Learning from Human Feedback (RLHF)~\cite{Ouyang2022TrainingLM}, such as Proximal Policy Optimization (PPO)~\cite{schulman2017proximal}, which optimizes model outputs based on reward signals produced by a reward model trained from human preference data. While effective, PPO's complex training pipeline can be a barrier to scalability.
Direct Preference Optimization (DPO)~\cite{Rafailov2023DirectPO} simplifies the training process by directly optimizing a classification objective over human preference pairs.
Models like DeepSeek-R1~\cite{DeepSeekAI2025DeepSeekR1IR} adopt the Reinforcement Learning with Verifiable Rewards (RLVR) paradigm, where verifiable outcomes from tasks are used as reward signals. However, these approaches still rely on external supervision or task-specific setups. Furthermore, as LLMs increasingly approach or exceed human expert performance, concerns of the long-term feasibility of obtaining high-quality human-labeled datasets have emerged~\cite{Yue2025DoesRL, Hughes2024OpenEndednessIE}. 

\begin{figure*}[t]
  \centering
  \includegraphics[width=0.92\linewidth]{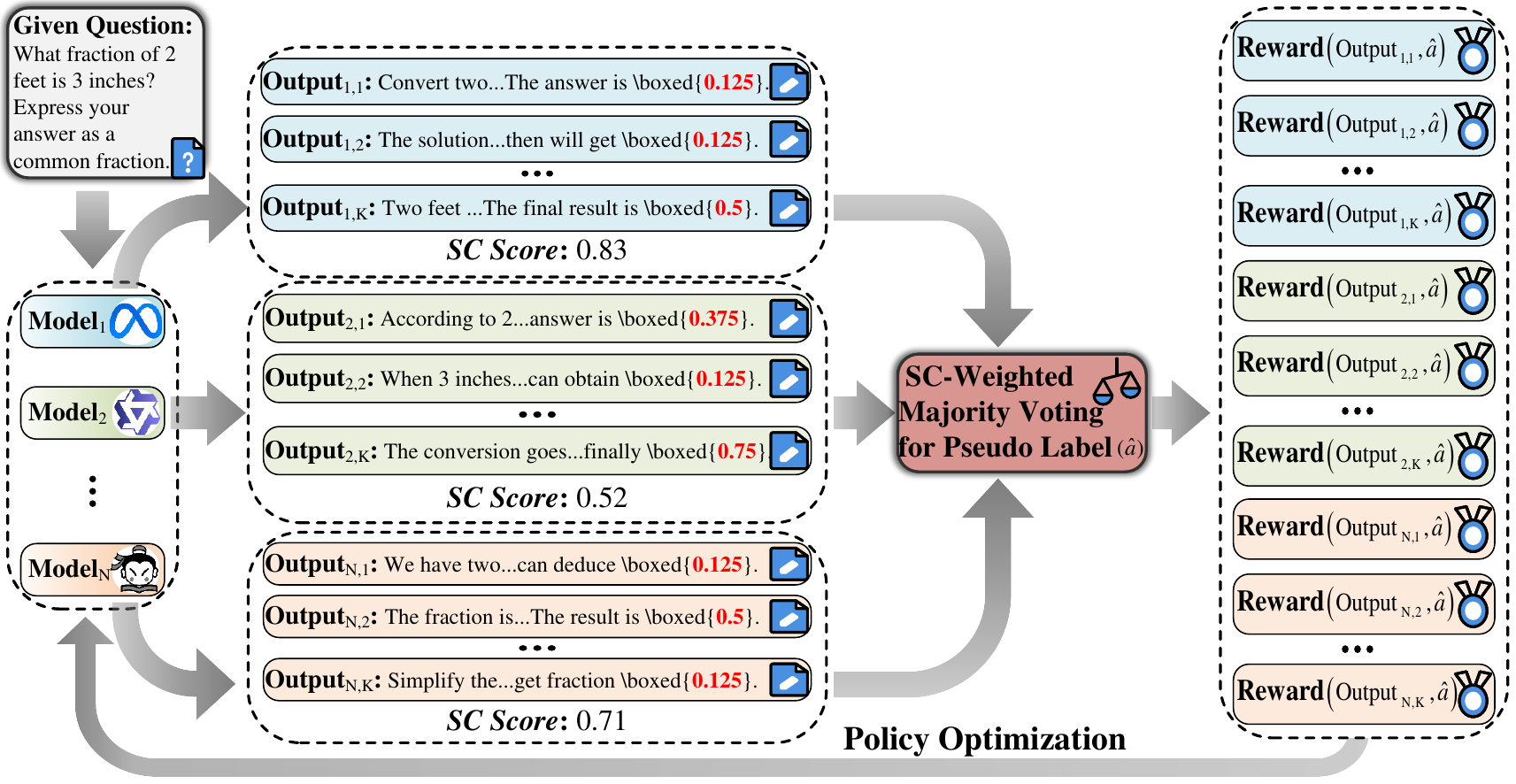}
  \caption{Overview of the RLCCF framework. Multiple models independently generate candidate outputs for a given question. SC-Weighted Majority Voting generates a high-quality pseudo-label, which acts as a shared supervision signal for all models.}
  \label{fig:framework}
\end{figure*}
\subsection{Self-Feedback RL}

The self-feedback paradigm has emerged as a promising research direction for reducing dependence on external supervision\cite{Chen2024SelfPlayFC, Yuan2024SelfRewardingLM}. TTRL~\cite{Zuo2025TTRLTR} estimates pseudo-labels via majority voting over multiple model outputs, constructing an unsupervised reward signal. Intuitor~\cite{Zhao2025LearningTR} leverages the model’s Self-Certainty as a feedback signal to guide the learning process. EMPO~\cite{Zhang2025RightQI} improves output stability by minimizing the entropy of generated responses. Genius~\cite{xu2025genius} generates a global-aware response sequence by simulating future steps and evaluates the value of candidate steps using the foresight score. 
However, these methods fundamentally rely on a single model to generate reward signals, which makes them susceptible to overconfidence in incorrect outputs, reward hacking, and even training collapse~\cite{Shafayat2025CanLR, shumailov2024ai}. Differently, RLCCF exploits the complementarity and diversity of output distributions across multi models to 
enable robust, scalable, and supervision-free training for a spiral consistent co-evolution of a model collective.

\section{Method}

\subsection{Preliminaries}

Reinforcement learning methods for large language models can be broadly categorized based on their reward acquisition strategies. Two primary paradigms are \textit{Reinforcement Learning from Human Feedback} (RLHF) and \textit{Reinforcement Learning with Verifiable Rewards} (RLVR).

\begin{figure}[!t]
\centering
\includegraphics[width=\linewidth]{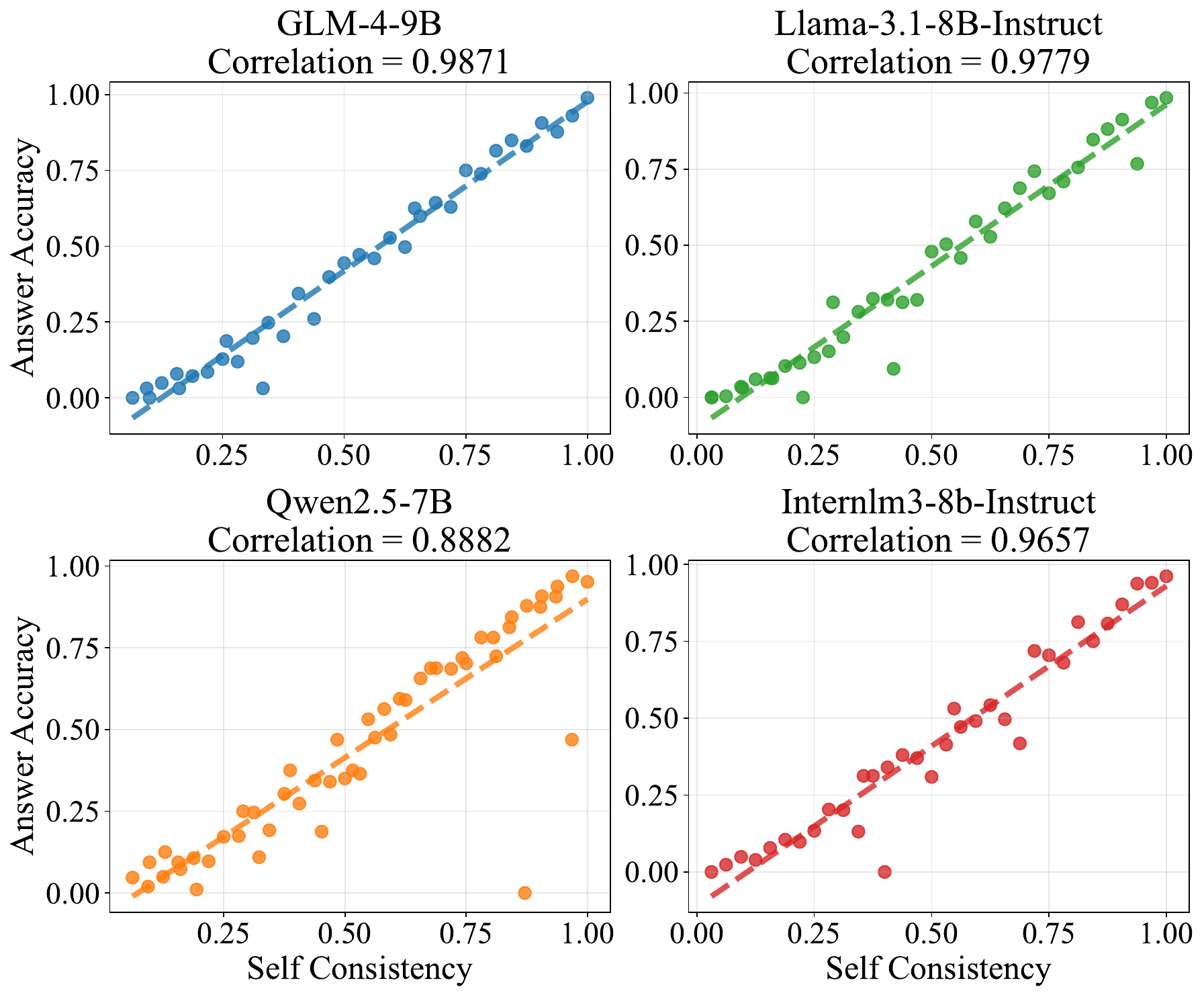}
\caption{Scatter plot of self-consistency versus answer accuracy for various models on the MATH-500 dataset, with correlation coefficients of 0.9871, 0.9779, 0.8882, and 0.9657 for GLM-4-9B, Llama-3.1-8B-Instruct, Qwen2.5-7B, and InternLM3-8B-Instruct, respectively.}
\label{fig:sc_acc}
\end{figure}

\textbf{RLHF} aims to align a policy model $\pi_\theta$ with human preferences, which are captured by a reward model $r_\phi$ trained on annotated comparison data. The RLHF objective is formulated as:
\begin{equation}
\label{eq:rlhf}
\max_{\pi_\theta}\; \mathbb{E}_{q \sim P(Q)} \left[ r_\phi(q, o) - \beta\, \mathrm{KL}\left(\pi_\theta(\cdot|q) \,\|\, \pi_{\text{ref}}(\cdot|q)\right) \right]
\end{equation}
where $q$ denotes the input query, $o \sim \pi_\theta(\cdot|q)$ is the generated output, $\pi_{\text{ref}}$ is a reference policy, and $\beta$ is a regularization coefficient that controls the divergence from the reference policy. The KL term serves to stabilize training by preventing large deviations from $\pi_{\text{ref}}$. 

A central challenge in RLHF lies in the construction of $r_\phi$, whose quality directly affects downstream model performance. Building an accurate reward model requires substantial human annotations and preference data~\cite{gao2023scaling}, significantly increasing both the complexity and cost of the training pipeline.

\textbf{RLVR} avoids the reward model training by leveraging automatically verifiable signals. In tasks with clear correctness criteria such as mathematical problem solving or code generation, rewards can be derived from a verification function $v(q, o)$ that evaluates output correctness. The corresponding optimization objective is:
\begin{equation}
\label{eq:rlvr}
\max_{\pi_\theta}\; \mathbb{E}_{q \sim P(Q)} \left[ v(q, o) - \beta\, \mathrm{KL}\left(\pi_\theta(\cdot|q) \,\|\, \pi_{\text{ref}}(\cdot|q)\right) \right]
\end{equation}
where $v(q, o)$ denotes the verifiable reward. While RLVR eliminates the burden of reward model training, its applicability is limited to tasks where verifier and human-labeled data are available\cite{Liu2025UnderstandingRT}.

\subsection{Reinforcement Learning from Coevolutionary Collective Feedback}

Multi-agent systems have demonstrated significant advantages over single-agent systems in collective decision-making and cooperative reasoning tasks~\cite{chen2025symbolic,lu2023routing}. Prior research has shown that aggregating diverse and independent opinions often leads to superior decision quality compared to relying on a single agent~\cite{shnitzer2023large,chen2024routerdc}. Inspired by these works, we incorporate multi-agent paradigm into self-feedback reinforcement learning for LLMs and propose \textbf{Reinforcement Learning from Coevolutionary Collective Feedback (RLCCF)}. RLCCF orchestrates a group of heterogeneous LLMs to enhance the accuracy and robustness of pseudo-label estimation in unsupervised settings.

\subsubsection{Pseudo-labels Estimation}
RLCCF estimates pseudo-labels for each input query \(q\) through multi-model collaborative voting mechanism. Let \(\mathcal{M} = \{M_1, M_2, \dots, M_N\}\) denote a set of \(N\) distinct models. Each model \(M_n\) independently generates \(K\) candidate responses \(\{o_{n,k}\}_{k=1}^K\) for the query \(q\). An answer extraction function is applied to filter invalid outputs. All of the valid answers are collected into the set \(\mathcal{O} = \{o_{n,k} \mid o_{n,k} \neq \varnothing\}\).

To assess the reliability of each model’s responses, we use SC score~\cite{Huang2022LargeLM} as a measure of model agreement with itself. SC is defined as the frequency of its most common answer. We performed an empirical study to validate the effectiveness of SC for identifying more accurate models in multi-model voting settings. Figure~\ref{fig:sc_acc} illustrates the relationship between self-consistency and answer accuracy on MATH-500 dataset. The results demonstrate a strong positive correlation between SC and answer accuracy, with a correlation coefficient of 0.9871 for GLM-4-9B, 0.9779 for Llama-3.1-8B-Instruct, 0.8882 for Qwen2.5-7B, and 0.9657 for InternLM3-8B-Instruct. This indicates that models with higher SC tend to provide more accurate responses, suggesting SC an effective criterion for selecting reliable models in collaborative voting.

Based on the observation, we employ a self-consistency weighted majority vote to determine the final pseudo-label \(\hat{a}\). Specifically, for each model \(M_n\), we compute \(SC_n\), which is then used as a weight in the voting process. The pseudo-label \(\hat{a}\) is selected as the answer that receives the highest aggregated SC-weighted vote across all models:
\begin{equation}
\hat{a} = \arg\max_a \sum_{n=1}^N \sum_{k=1}^{K}\text{SC}_n \cdot \mathbb{I}[a = o_{n,k}]
\end{equation}
where \(\mathbb{I}[\cdot]\) denotes the indicator function, which is 1 if \(a\) is equal to \(o_{n,k}\), and 0 otherwise.

This voting mechanism allows RLCCF to generate robust and high-confidence pseudo-labels by integrating both answer frequency and model self-consistency, ensuring that the final label reflects the consensus of the most reliable models.

\subsubsection{Policy Optimization}
To encourage model outputs to align with the collective consensus, we define the reward for each candidate output \(o_{n,k}\) as:
\begin{equation}
\label{eq:reward-binary}
r_{n,k} =
\begin{cases}
1, & \text{if}~o_{i,k} = \hat{a} \\
0, & \text{otherwise}
\end{cases}
\end{equation}

The overall objective of RLCCF is to maximize collective consistency. We adopt the Group Relative Policy Optimization (GRPO)~\cite{guo2025deepseek} framework for policy optimization, but introduce our group reward mechanism derived from our coevolutionary collective feedback. To ensure scalability and computational tractability, each model adopts an independent objective. Specifically, each model \(M_n\) optimizes its policy parameters \(\pi_{\theta_n}\) according to its own sampled outputs \(\{o_{n,k}\}_{k=1}^K\) and the group-voted pseudo-label \(\hat{a}\). The model \(M_n\) is optimized by maximizing:
\begin{equation}
\label{eq:grpo-main}
\max_{\pi_{\theta_n}}~ \mathbb{E}_{q \sim P(Q)} \left[ \mathcal{J}^{}_{n}(q) \right]
\end{equation}
where the optimization objective for each model \(M_n\) is defined as:
\begin{align}
\label{eq:grpo-loss}
\mathcal{J}^{}_{n}(q) = \frac{1}{K} \sum_{k=1}^K \widehat{A}_{n,k} - \beta\, \mathrm{KL}\big(\pi_{\theta_n}(\cdot|q) \,\|\, \pi_{\text{ref}_n}(\cdot|q)\big)
\end{align}
where \(\widehat{A}_{n,k}\) denotes the advantage for the \(k\)-th output, computed as:
\begin{equation}
\widehat{A}_{n,k} = \min\!\left(\rho_{n,k} A_{n,k},~ \operatorname{clip}(\rho_{n,k}, 1-\epsilon, 1+\epsilon)\,A_{n,k}\right)
\end{equation}

\begin{equation}
\rho_{n,k} = \frac{\pi_{\theta_n}(o_{n,k} \!\mid q)}{\pi_{\theta_n^{\text{old}}}(o_{n,k} \!\mid q)}
\end{equation} 
where \(\epsilon\) is a small constant controlling the clipping range of the advantage. The normalized advantage \(\widehat{A}_{n,k}\) within each group is given by:
\begin{equation}
A_{n,k} = \frac{r_{n,k} - \mathrm{mean}(r_{n,1}, \dots, r_{n,K})}{\mathrm{std}(r_{n,1}, \dots, r_{n,K})}
\end{equation}
where \(r_{n,k}\) is the reward for output \(o_{n,k}\).

This collaborative, unsupervised learning framework enables robust and cost-efficient improvement of LLMs by maximizing the agreement within a diverse group of models, even in the absence of labeled reference answers.

\subsection{Why Does Collaboration Help?}

\subsubsection{Bias Reduction}

To theoretically analyze the effectiveness of our multi-model collaborative approach, we model how aggregating outputs from a diverse ensemble of LLMs enhances the quality of supervision signals.

We begin by modeling the output distribution of each individual LLM. Let's consider an ensemble of $N$ models, with each model $n$ generating $K$ candidate answers for a given problem. 
Similar to \cite{Guha2024SmoothieLF}, we assume the output embedding $X_{n,k}$ from model $n$ for the $k$-th sample follows a normal distribution:
\[
X_{n,k} \sim \mathcal{N}(\text{GT} + \epsilon_n, \sigma_n^2)
\]
where $\text{GT}$ is the ground-truth answer embedding, $\epsilon_n$ is the model-specific bias, and $\sigma_n^2$ is the model's internal variance. We assume the biases $\epsilon_n$ are independent random variables with a zero mean across the ensemble, i.e., $\mathbb{E}[\epsilon_n]=0$. This assumption is reasonable given the models in our ensemble come from different vendors, utilize varied architectures, and be trained on distinct datasets. Such diversity helps ensure that model biases are not systematically correlated.

The process of aggregating outputs from all $N$ models can be viewed as sampling from a aggregated distribution. 
The distribution of the sample mean from a collection of independent random variables (in our case, the outputs from different models) will tend to a normal distribution. The aggregated output, $X = \frac{1}{NK}\sum_{n=1}^{N}\sum_{k=1}^{K}X_{n,k}$, will thus follow a new normal distribution:
\[
X \sim \mathcal{N}(\mathbb{E}[X], \text{Var}(X))
\]
We can then compute the parameters of this distribution:
\begin{equation*}
\begin{aligned}
\mathbb{E}[X] &= \mathbb{E}\left[\frac{1}{NK} \sum_{n=1}^{N}\sum_{k=1}^{K} X_{n,k}\right] \\
&= \frac{1}{N}\sum_{n=1}^{N}(\text{GT} + \epsilon_n) = \text{GT} + \frac{1}{N}\sum_{n=1}^{N}\epsilon_n
\end{aligned}
\end{equation*}
Since the model biases $\epsilon_n$ are independent with a zero mean, the average bias term $\frac{1}{N}\sum_{n=1}^{N}\epsilon_n$ tends to zero as the number of models, $N$, increases. Consequently, the mean of the aggregated distribution converges to the ground truth:
\[
\mathbb{E}[X] \to \text{GT} \quad \text{as } N \to \infty
\]

A key property of the normal distribution is that its mean corresponds to its maximum probability density. In our context, this means the most probable aggregated output is the ground truth. This theoretical result provides a strong foundation for our \textbf{group majority-voting mechanism}, which empirically seeks the most frequent (i.e., most probable) answer among all samples. By performing this vote, we are effectively selecting the value with the highest probability density from the aggregated distribution, which is the ground truth answer. Therefore, multi-model aggregation dramatically increases the probability of correct majority vote result, leading to high-quality and robust pseudo-labels for RL.

\subsubsection{Model Complementarity}

\begin{figure}[t]
\centering
\includegraphics[width=\linewidth]{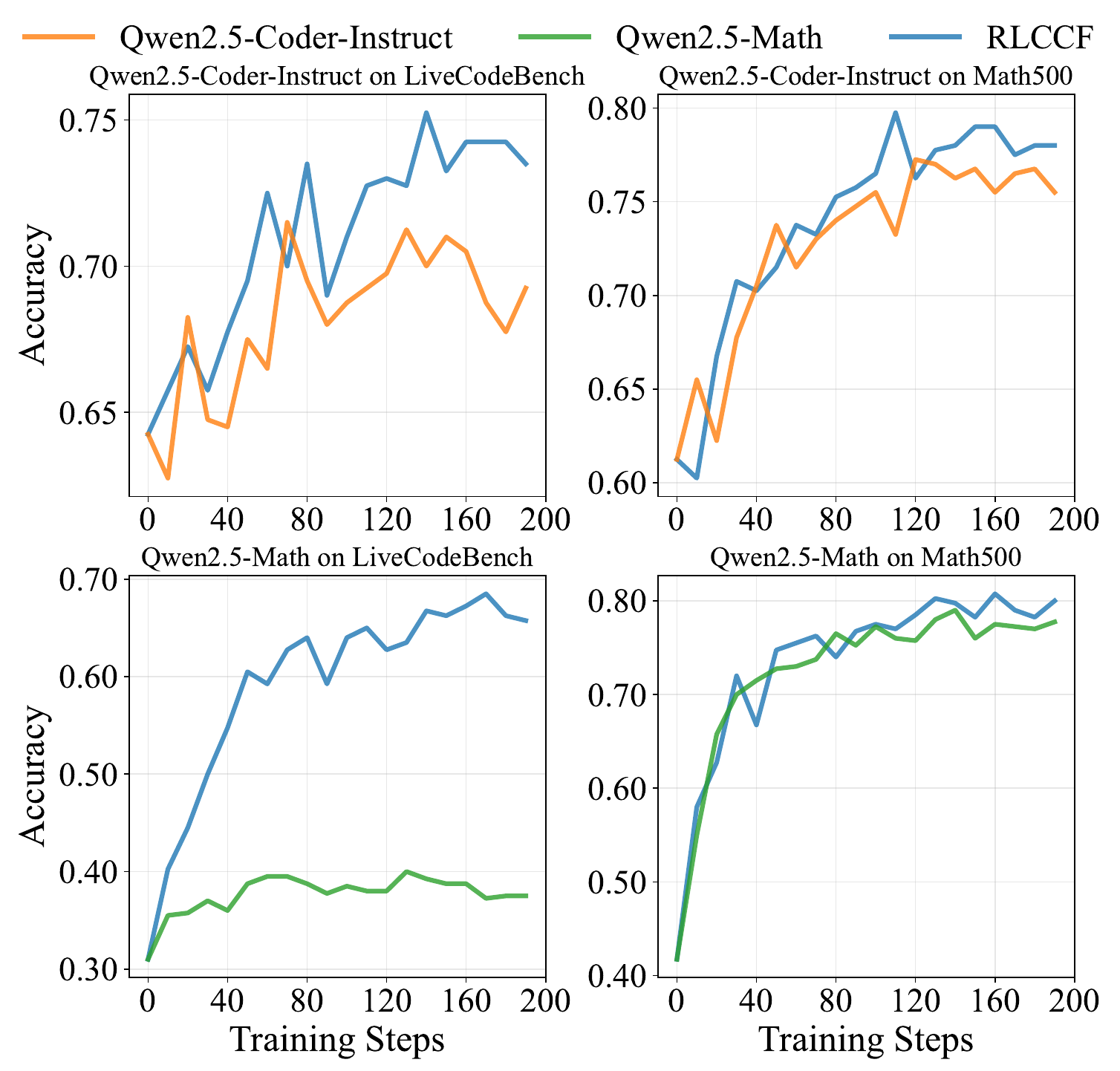}
\caption{Accuracy curves comparing individual models with the collaborative group (RLCCF) across a mixed LiveCodeBench and MATH-500 dataset. 
}
\label{fig:model_complementarity}
\end{figure}

To investigate the synergistic effects of multi-model collaboration, we analyze RLCCF from the perspective of model complementarity. We constructed a mixed-domain dataset, comprising 100 problems from LiveCodeBench-Execution (code) and 100 from MATH-500 (mathematics), to evaluate the performance of two domain-specialized models: Qwen2.5-Math-7B and Qwen2.5-Coder-7B-Instruct.

For the training, we allocated 64 samples per problem for reward estimation and 16 samples for policy updates. As depicted in Figure~\ref{fig:model_complementarity}, the individual models exhibit a clear pattern: they perform well within their respective specialized domains but struggle with tasks outside their expertise. In sharp contrast, the collaborative group, by leveraging the complementary strengths of the math-oriented and code-oriented models, demonstrates consistent performance improvement during training. This synergistic effect leads to significantly better overall performance across both domains, highlighting that RLCCF not only reduces collective bias but also fosters cross-domain learning by combining specialized knowledge.

\section{Experiments}

\begin{table*}[t]
\centering
\resizebox{0.9\linewidth}{!}{
\begin{tabular}{llccccccccccc}
\toprule
\textbf{Model} & \textbf{Method} & \textbf{Supervision} & \textbf{Math-500} & \textbf{AMC} & \textbf{AIME} & \textbf{Olympiad} & \textbf{AVG} \\
\midrule
\multirow{5}{*}{Qwen2.5-7B}
& Base                 & -                    & 60.04          & 32.91          & 8.13           & 26.58          & 31.92          \\
& Intuitor             & Self                 & 71.83          & 42.09          & 10.00          & 34.13          & 39.40          \\
& EMPO                 & Self                 & 73.00          & 41.72          & \textbf{12.40} & 34.97          & 40.52          \\
& TTRL                 & Self                 & \textbf{73.08} & 41.75          & 12.19          & 34.23          & 40.31          \\
& RLCCF                & Group                & \textbf{73.08} & \textbf{42.43} & 11.46          & \textbf{35.41} & \textbf{40.60} \\
\midrule
\multirow{5}{*}{GLM-4-9B}
& Base                 & -                    & 62.02          & 30.35          & 9.79           & 29.80          & 32.99          \\
& Intuitor              & Self                 & 62.14          & 31.25          & 9.90           & 30.12          & 33.35          \\
& EMPO                 & Self                 & 64.63          & 36.71          & \textbf{12.60} & 28.86          & 35.70          \\
& TTRL                 & Self                 & 67.47          & 35.99          & 11.35          & 33.04          & 36.96          \\
& RLCCF                & Group                & \textbf{69.43} & \textbf{39.27} & 11.15          & \textbf{34.05} & \textbf{38.48} \\
\midrule
\multirow{5}{*}{InternLM3-8b-Instruct}
& Base                 & -                    & 65.44          & 33.36          & 8.13           & 31.12          & 34.51          \\
& Intuitor              & Self                 & 65.91          & 34.94          & 8.02           & 6.67           & 28.89          \\
& EMPO                 & Self                 & 69.53          & \textbf{39.87} & 11.67          & 35.37          & 39.11          \\
& TTRL                 & Self                 & 70.84          & 39.04          & 12.71          & \textbf{36.00} & 39.65          \\
& RLCCF                & Group                & \textbf{71.24} & 39.01          & \textbf{13.23} & 35.70          & \textbf{39.80} \\
\midrule
\multirow{5}{*}{LLaMA-3.1-8B-Instruct}
& Base                 & -                    & 48.92          & 22.21          & 6.25           & 17.32          & 23.68          \\
& Intuitor              & Self                 & 3.24           & 4.22           & 0.00           & 1.46           & 2.23           \\
& EMPO                 & Self                 & \textbf{51.94} & 22.44          & 7.92           & \textbf{18.05} & 25.09          \\
& TTRL                 & Self                 & 50.27          & 22.82          & 6.46           & 16.46          & 24.00          \\
& RLCCF                & Group                & 50.12          & \textbf{25.49} & \textbf{8.44}  & 17.98          & \textbf{25.51} \\
\midrule
\midrule
\multirow{5}{*}{Group Majority Vote}
& Base                 & -                    & 81.20          & 50.60          & 20.00          & 43.01          & 48.70          \\
& Intuitor              & Self                 & 81.40          & 49.40          & 13.33          & 35.27          & 44.85          \\
& EMPO                 & Self                 & 82.00          & 51.81          & 20.00          & 42.56          & 49.09          \\
& TTRL                 & Self                 & 82.00          & 48.19          & 16.67          & 45.39          & 48.06          \\
& RLCCF                & Group                & \textbf{82.20} & \textbf{54.22} & \textbf{20.00} & \textbf{47.17} & \textbf{50.90} \\

\bottomrule
\end{tabular}
}
\caption{Performance comparison of various models and methods across mathematical reasoning benchmarks. The best results within each model are highlighted in bold. Note that due to collapse encountered with Intuitor  across multiple experiments, we report results from its best-performing checkpoint. All methods use the same computational budget, with 64 samples for reward estimation and 16 samples during policy optimization.}
\label{tab:multi-model-results}
\end{table*}

\subsection{Baselines and Setup}
\label{Baselines and Setup}

We conduct our experiments under the VERL framework~\cite{Sheng2024HybridFlowAF} with four representative LLMs from different developers: InternLM3-8B-Instruct~\cite{cai2024internlm2}, GLM-4-9B~\cite{glm2024chatglm}, Qwen2.5-7B~\cite{Yang2024Qwen25TR}, and LLaMA3.1-8B-Instruct~\cite{Dubey2024TheL3}. To construct a training corpus tailored for mathematical reasoning, we randomly sample 100 problems from each of the seven subcategories in the MATH training set~\cite{Hendrycks2021MeasuringMP}, yielding a dataset denoted as MATH-700.

To evaluate the effectiveness of our proposed approach, we compare it against three strong self-feedback RL baselines: TTRL~\cite{Zuo2025TTRLTR}, EMPO~\cite{Zhang2025RightQI}, and Intuitor~\cite{Zhao2025LearningTR}. Following the original setup, Intuitor generates 16 candidate outputs per problem. For TTRL and EMPO, we adopt the same sampling configuration as TTRL, producing 64 candidate outputs per problem for reward estimation while retaining only 16 samples for policy updates to ensure a fair comparison with Intuitor in terms of optimization cost.

For our proposed RLCCF framework, each model in the group independently generates 16 candidate outputs, yielding a total of 64 outputs used for group-based reward estimation. During policy optimization, each model updates its parameters using only its own 16 outputs. This design ensures consistency in both sampling scale and optimization logic across all methods, enabling fair and controlled comparisons. All methods are trained on MATH-700 for 3 epochs with a batch size of 8, under identical hypermainparameter configurations.

\subsection{Evaluation}
We evaluate model performance on four math reasoning benchmarks of varying difficulty, including AIME 2024~\cite{li2024numinamath}, AMC~\cite{li2024numinamath}, MATH-500~\cite{hendrycks2021measuring}, and OlympiadBench~\cite{he2024olympiadbench}, with difficulty roughly ordered as AIME $>$ OlympiadBench $>$ AMC $>$ MATH-500.

During evaluation, each model is prompted to reason step-by-step, with the final answer required to be within \verb|\boxed{}|. For each problem, we sample 32 responses and report the average accuracy across the sampled outputs. For the group majority vote evaluation, we apply each baseline method to train the four models independently and aggregate their outputs for voting with 32 samples. The evaluation protocol is based on the official TTRL implementation under VERL framework to ensure fairness and consistency across all methods. All experiments are conducted on a cluster equipped with 2 * 8 NVIDIA A800 80GB GPUs.

\subsection{Main Results and Analysis}

Table \ref{tab:multi-model-results} demonstrates the consistent and substantial improvements of our proposed RLCCF framework across various language models. RLCCF achieves average accuracies of 40.60\% on Qwen2.5-7B, 38.48\% on GLM-4-9B, 39.80\% on InternLM3-8b-Instruct, and 25.51\% on LLaMA-3.1-8B-Instruct, representing relative improvements over the baseline of 27.20\%, 16.63\%, 15.31\%, and 7.74\%, respectively. Moreover, RLCCF surpasses existing self-feedback methods, outperforming Intuitor  by 38.99\%, EMPO by 2.82\%, and TTRL by 2.45\%. A key observation is the stability of RLCCF, while Intuitor  struggled with instability and occasional performance collapse during multi-model training, RLCCF provides robust and stable gains. 
Notably, while most single-model self-feedback methods either fail to improve or even harm the group's majority-vote accuracy, our RLCCF framework delivers a substantial breakthrough, boosting the group's majority vote accuracy from 48.70 \% to 50.90\%. This indicates that our method expands the capability boundary of the entire model group. These results underline the effectiveness and robustness of RLCCF in enhancing mathematical reasoning capabilities across diverse model architectures and benchmarks.

\subsubsection{Reward Accuracy and Label Accuracy}

\begin{figure}[t]
\centering
\includegraphics[width=\linewidth]{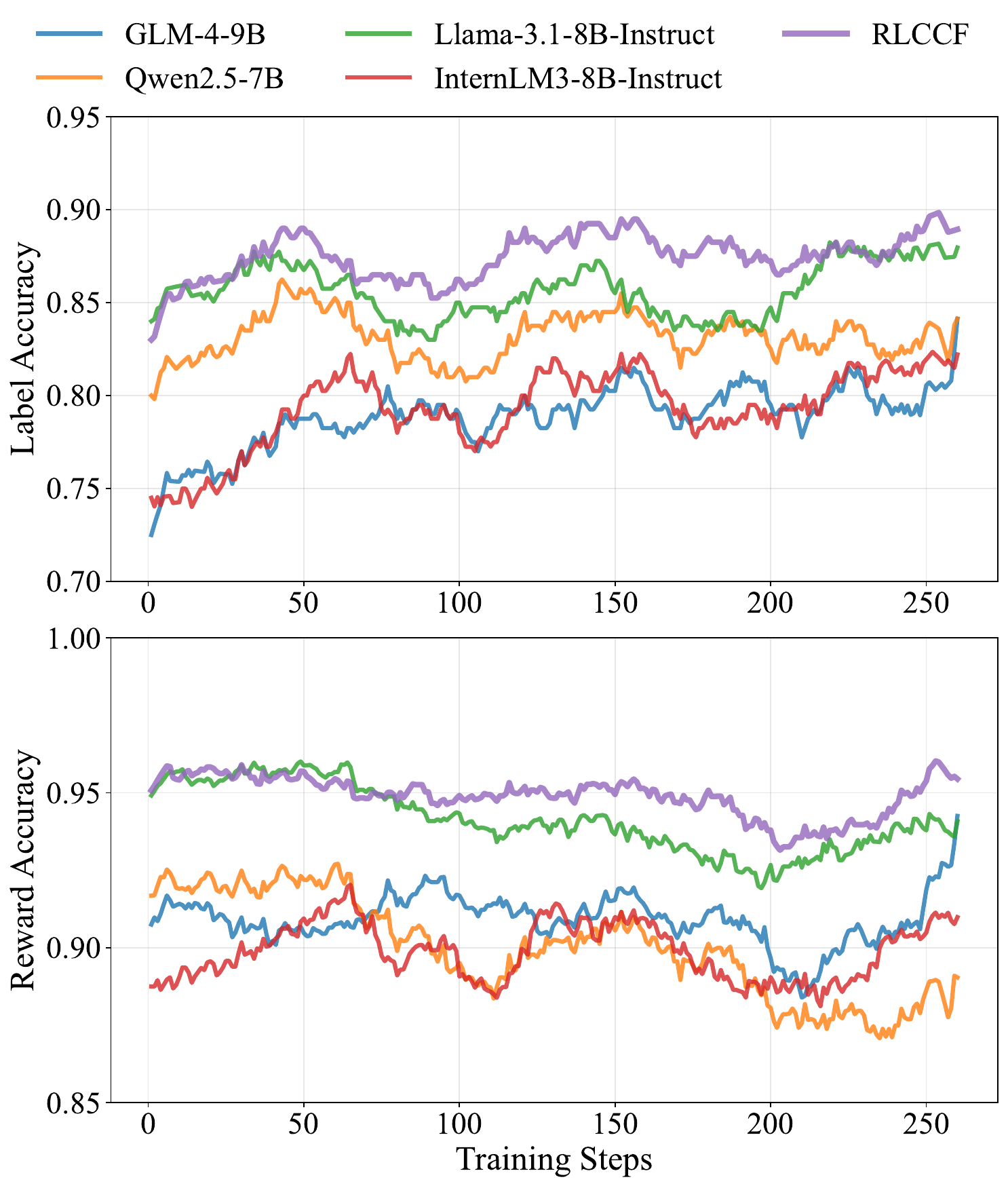}
\caption{Comparison of reward and label accuracy between RLCCF and the TTRL baseline. A moving average over 50 training steps is applied for clarity.}
\vspace{-10pt}
\label{fig:reward_label_acc}
\end{figure}

To evaluate the effectiveness of our collaborative training paradigm, we compare two key metrics: reward accuracy and label accuracy, between our group-based method and the single-model baseline, TTRL. As shown in Figure~\ref{fig:reward_label_acc}, we control the total voting budget to be 64 for both settings. 

The label accuracy steadily increases over the process of training, indicating that the models are progressively improving their mathematical reasoning ability
Notably, our group-based method consistently outperforms the individual models on both metrics with a more stable training process. It demonstrates that by harnessing the collective intelligence of diverse models, our collaborative evolution framework generates more accurate pseudo-labels and more reliable reward signals, enabling more effective policy updates.

\subsubsection{Emergence of Higher Collective Consistency}

\begin{figure}[t]
\centering
\includegraphics[width=\linewidth]{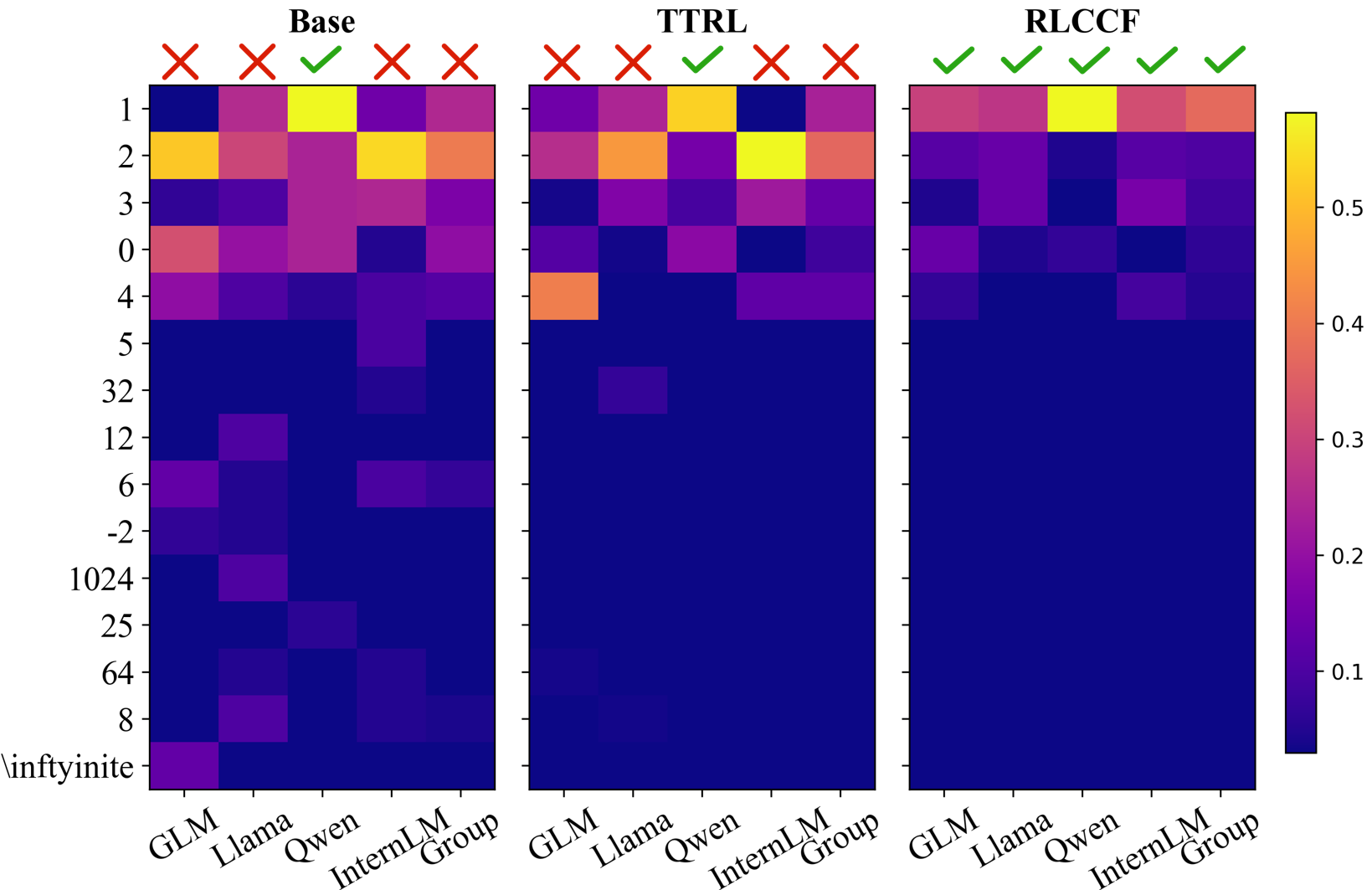}
\caption{A representative case study illustrating the answer distribution heatmaps for individual models and the collective group under different training strategies. The ground-truth answer for the example is \textbf{1}.}
\label{fig:Collective Consistence}
\end{figure}

As shown in Figure \ref{fig:Collective Consistence}, a higher degree of collective consistency is observed in the model group after RLCCF training. In this representative case, while both TTRL and RLCCF improve the consistency of individual model responses compared to the base models, TTRL-trained models exhibit different answer distribution peaks, which leads to an incorrect group majority vote. In contrast, the models trained with RLCCF converge to a more consistent peak, resulting in a more accurate group majority vote.

\subsubsection{Ablation}

\begin{table}[t]
\setlength{\tabcolsep}{1mm}
\centering
\begin{tabular}{lccc}
\toprule
\textbf{Model} & \textbf{Simple} & \textbf{SC Weighted} & \textbf{Gain} \\
\midrule
Qwen2.5-7B & 39.74 & 40.60 & +0.86 \\
GLM-4-9B & 37.44 & 38.48 & +1.04 \\
InternLM3-8B-Instruct & 39.54 & 39.80 & +0.26 \\
LLaMA-3.1-8B-Instruct & 25.09 & 25.51 & +0.42 \\
Group Majority Vote & 49.61 & 50.90 & +1.29 \\
\bottomrule
\end{tabular}
\caption{Comparison between simple majority voting and SC weighted voting across different models. Gain indicates the performance improvement from SC weighting.}
\vspace{-10pt}
\label{tab:sc_weighted_gain}
\end{table}

We conduct an ablation study on the voting strategy employed during training in the RLCCF framework. As shown in Table~\ref{tab:sc_weighted_gain}, incorporating self-consistency (SC) weighted voting consistently enhances performance across all individual models when compared to simple majority voting. Notably, the group-level majority vote also exhibits a significant gain, suggesting that SC scores serve as effective confidence indicators for guiding the ensemble toward more reliable pseudo-labels.








\section{Conclusion}


In this work, we propose Reinforcement Learning from Coevolutionary Collective Feedback (RLCCF) to overcome the limitations of single-model self-feedback. RLCCF uses a dynamically weighted majority vote, based on each model's Self-Consistency (SC) score, to generate robust pseudo-labels from collective knowledge. Extensive experiments on four open-source LLMs across four mathematical benchmarks showed that RLCCF significantly improves individual model performance. Our findings further demonstrate that RLCCF not only enhances individual models but also advances the collective capability of the entire ensemble.

\bibliography{aaai2026}

\end{document}